\documentclass[journal]{IEEEtai}
\usepackage{amsmath,amsfonts}
\usepackage{algorithmic}
\usepackage{array}
\usepackage[caption=false,font=normalsize,labelfont=sf,textfont=sf]{subfig}
\usepackage{textcomp}
\usepackage{stfloats}
\usepackage{url}
\usepackage{verbatim}
\usepackage{graphicx}
\usepackage[utf8]{inputenc}
\usepackage[T1]{fontenc}
\usepackage{endnotes}
\usepackage{tabularx}

\usepackage{siunitx}
\usepackage{graphicx}
\usepackage{float}
\usepackage{subcaption}
\usepackage{microtype}
\usepackage{multirow}
\usepackage{nicefrac}
\usepackage{amsmath, amsfonts}

\usepackage[dvipsnames]{xcolor}
\definecolor{DarkSlateGray}{RGB}{47,79,79}
\definecolor{DarkViolet}{RGB}{148,0,211}
\definecolor{DarkOrange}{RGB}{255,140,0}
\definecolor{ReddishOrange}{RGB}{220,80,40}

\usepackage{tcolorbox}
\tcbuselibrary{most, listings}

\usepackage{listings}
\usepackage{xcolor}
\usepackage[table]{xcolor}
\usepackage[numbers]{natbib}

\lstdefinelanguage{json}{
  showstringspaces=false,
  breaklines=true,
  morestring=[b]",
  stringstyle=\color{black},
  literate=
   *{:}{{:}}1
    {,}{{,}}1
    {\{}{{\{}}1
    {\}}{{\}}}1
    {[}{{[}}1
    {]}{{]}}1
}

\lstdefinestyle{jsonstyle}{
  language=json,
  basicstyle=\ttfamily\small,
  columns=fullflexible,
  breaklines=true,
  showstringspaces=false
}

\lstdefinestyle{jsonstyle}{
  language=json,
  basicstyle=\ttfamily\small,
  showstringspaces=false,
  breaklines=true,
  columns=fullflexible
}

\newtcblisting{jsonlisting}{
  listing only,
  breakable,
  colback=gray!4,
  colframe=black!30,
  boxrule=0.3pt,
  arc=1pt,
  left=1mm, right=1mm, top=1mm, bottom=1mm,
  listing options={style=jsonstyle}
}

\usepackage{url}
\usepackage{booktabs}
\usepackage{orcidlink}
\usepackage{hyperref}

\def\BibTeX{{\rm B\kern-.05em{\sc i\kern-.025em b}\kern-.08em
    T\kern-.1667em\lower.7ex\hbox{E}\kern-.125emX}}
\usepackage{balance}

\begin{document}


\title{RGPO: Ranking-Guided Preference Optimization for Reliable Clinical Reasoning}

\author{
Chia-Hsuan Hsu$^{*}$,
Jun-En Ding$^{*}$,
Hsin-Ling Hsu,
Chih-Ho Hsu,
Li-Hung Yao,
Chun-Chieh Liao,
Feng Liu,
and Fang-Ming Hung
\thanks{$^{*}$These authors contributed equally to this work.}
\thanks{Corresponding authors: F. Liu and F.-M. Hung. Emails: fliu22@stevens.edu, philip@mail.femh.org.tw.}
\thanks{C.-H. Hsu is with the Department of Computer Science and Information Engineering,
National Taiwan University of Science and Technology, Taipei City, Taiwan.}
\thanks{J.-E. Ding is with the Department of Systems Engineering, Stevens Institute of Technology,
Hoboken, NJ, USA, and also with Far Eastern Memorial Hospital, New Taipei City, Taiwan.}
\thanks{H.-L. Hsu is with the Department of Management Information Systems, National Chengchi University,
Taipei City, Taiwan, and also with Far Eastern Memorial Hospital, New Taipei City, Taiwan.}
\thanks{C.-H. Hsu is with Far Eastern Memorial Hospital, New Taipei City, Taiwan.}
\thanks{L.-H. Yao is with the AI Research Center, Agaruda System, Taipei City, Taiwan.}
\thanks{C.-C. Liao is with the Department of Computer Science,
Stevens Institute of Technology, Hoboken, NJ, USA.}
\thanks{F. Liu is with the Department of Systems Engineering,
Stevens Institute of Technology, Hoboken, NJ, USA.}
\thanks{F.-M. Hung is with Far Eastern Memorial Hospital, New Taipei City, Taiwan,
and also with the Smart Healthcare Interdisciplinary College,
National Taipei University of Nursing and Health Sciences, Taipei City, Taiwan.}
}


\maketitle

\begin{abstract}
Medical question answering in healthcare systems requires advanced reasoning that integrates domain knowledge with clinical workflows. However, existing large language models (LLMs) often generate reasoning chains that lack factual accuracy and clinical reliability. We propose Ranking-Guided Preference Optimization (RGPO) for Reliable Clinical Reasoning , a framework that combines reinforcement learning with preference-driven reasoning refinement to enhance clinical chain-of-thought performance in LLM-based healthcare systems. RGPO employs task-adaptive reasoning templates aligned with clinical protocols and a probabilistic evaluation mechanism that integrates with healthcare information systems, including Electronic Health Records (EHRs) and clinical documentation, while automatically identifying and correcting low-quality reasoning chains. Unlike traditional pairwise preference methods, RGPO introduces groupwise ranking optimization based on the Bradley-Terry model with Kullback–Leibler (KL)-divergence regularization for stable clinical deployment. The framework addresses key challenges, including multimodal data integration, clinical validation, and real-world interoperability. Experiments on PubMedQA, MedQA-USMLE, and real-world validation at Far Eastern Memorial Hospital (FEMH) demonstrate consistent improvements over strong baselines. Our 2B-parameter model outperforms larger 7B–20B models, including medical-specialized variants, while maintaining computational efficiency for healthcare deployment. These results demonstrate that combining preference optimization with quality-driven refinement provides a scalable, clinically validated approach to building reliable medical LLMs for practical healthcare applications.
\end{abstract}

\begin{IEEEImpStatement}
This work presents Ranking-Guided Preference Optimization (RGPO), a reinforcement optimization framework that significantly improves the medical reasoning capabilities of Large Language Models (LLMs). Current medical LLMs often struggle with multi-step diagnostic logic, factual reliability, and safety alignment, limiting their use in real clinical workflows. RGPO uses ranked preference feedback to guide models toward more accurate, interpretable, and clinically consistent reasoning. Experiments across multiple medical benchmark datasets demonstrate clear performance gains over the original base models, indicating that RGPO effectively enhances reasoning quality in diverse clinical scenarios. These advances strengthen the safety and trustworthiness of medical AI systems, reduce hallucination risks in high-stakes applications, and provide a scalable strategy for aligning LLMs with clinical decision needs. The proposed approach supports the development of more dependable AI assistants for clinicians, medical training, and patient-facing healthcare services.
\end{IEEEImpStatement}


\section{Introduction}
Recent advances in large language models (LLMs) signal significant promise for medical applications, especially in clinical reasoning and diagnostic support systems~\cite{singhal2025toward, thirunavukarasu2023large, gilson2023does}. However, deploying AI in medicine requires exceptional reliability, making factual accuracy and thorough coverage of the data paramount. Medical reasoning introduces challenges distinct from those in general domains: clinical decisions require a comprehensive consideration of relevant factors, strict factual accuracy, and the avoidance of redundant information that could obscure critical insights.


Current approaches to medical reasoning primarily rely on chain-of-thought (CoT) prompting techniques~\cite{wei2022chain, kojima2022large, wang2022self}, which have shown effectiveness in improving reasoning capabilities across various domains. However, these methods typically generate single reasoning chains without systematic quality assessment or iterative refinement mechanisms. Moreover, existing evaluation frameworks often rely on coarse-grained binary evaluation or preference signals that fail to capture the nuanced quality required for medical reasoning~\cite{liang2022holistic, pesaran2025lpoi, white2024rating} or require extensive human annotation, limiting their suitability for automated quality control where domain expertise is essential.

Recent work in preference learning and reinforcement learning from human feedback (RLHF)~\cite{ouyang2022training, christiano2017deep, rafailov2023direct, yuan2023rrhf} shows promise in aligning AI systems with human preferences. Direct Preference Optimization (DPO) \cite{rafailov2023direct} has shown promising results by learning from preference data. However, these methods typically rely on pairwise comparisons, which fail to capture all the ranking information in human preference datasets. Many real-world scenarios involve humans ranking multiple candidates, not just providing binary choices. For example, when they evaluate reasoning chains, mathematical solutions, or creative content, evaluators assign rankings that reflect nuanced quality differences. Reducing this rich data to pairwise comparisons can lose valuable training signals and lead to suboptimal solutions.

Specifically, these approaches face significant challenges in medical domains. First, the requirement for extensive expert annotation is prohibitively expensive~\cite{hendrycks2021measuring, han2024medsafetybench}. Additionally, preference judgments often lack the granularity needed for medical reasoning assessment~\cite{geva2019we, wu2024medjourney}. Furthermore, generic preference models fail to capture the domain-specific quality dimensions that are critical in healthcare applications~\cite{nori2023capabilities}.


To address these limitations, we introduce \textbf{Ranking-Guided Preference Optimization for
Reliable Clinical Reasoning (RGPO)}, a novel framework for improving the quality and reliability of medical reasoning. The proposed framework integrates a probabilistic refinement module for reasoning quality assessment with preference-based ranking optimization grounded in the Bradley--Terry model. RGPO combines domain-adaptive reasoning templates with automated quality evaluation and targeted revision mechanisms. Unlike existing methods that rely on single reasoning chains or costly human evaluation, our framework automatically generates multiple reasoning candidates, evaluates them using multi-dimensional probabilistic metrics, and performs targeted improvements. Our approach incorporates several key innovations:
\begin{itemize}
    \item A probabilistic framework for automated medical reasoning quality assessment that captures multi-dimensional evaluation criteria through conditional probability modeling;
    \item Domain-specific CoT enhancement templates that adapt reasoning structures to different medical contexts;
    \item A novel linear reward learning framework that provides a stable policy gradient for ranked preference optimization;
    \item A method for generating high-quality preference data suitable for RGPO fine-tuning, enabling continual improvement of medical reasoning capabilities.
\end{itemize}

\section{Related Work}

\subsection{Medical LLMs: Question Answering (QA), Diagnosis, and EHR}
Large language models (LLMs) have been used for medical tasks like knowledge-based question answering, diagnostic reasoning on clinical vignettes, and patient-specific reasoning over electronic health records (EHR) \citep{jin2019pubmedqa,jin2021disease,shi2024ehragent}. Domain-structured prompts and chain-of-thought (CoT) approaches consistently improve auditability and reduce unsupported claims by clearly separating decomposition and background from justification or diagnosis \citep{wei2022chain,singhal2025toward}. In patient-specific contexts, retrieval-augmented generation (RAG) and citation-style evidence tracing help ensure faithfulness to chart data and guidelines \cite{lewis2020rag,shi2024ehragent}. Recent work extends beyond answer accuracy, emphasizing multidimensional evaluation such as coverage of relevant findings and alignment with context, thereby better matching clinical expectations for completeness and correctness. In this work, we use domain-specific CoT templates (QA and diagnosis), automatically score rationales on coverage, factual accuracy, and redundancy with an LLM judge, and make light score-guided revisions.

\subsection{Reinforcement-Learned Medical LLMs}

Reinforcement learning from human feedback (RLHF) and related preference-based methods have become central to aligning LLM behavior beyond supervised fine-tuning (SFT)~\cite{christiano2017deep, ziegler2019fine, ouyang2022training}. In reasoning-heavy domains, variants that regularize to a reference policy while optimizing rewards derived from preference ordering or task-specific outcomes have shown improved stability and faithfulness.  

In the medical domain, reinforcement-learned LLMs have been applied to tasks such as clinical question answering and diagnostic reasoning~\cite{singhal2025toward, nori2023capabilities}. These methods typically employ reinforcement learning techniques such as policy optimization with preference feedback~\cite{christiano2017deep, ouyang2022training} to guide models toward multi-step clinical reasoning, integration of biomedical knowledge, and reliable elimination of alternative diagnoses.  

Prior studies further highlight persistent challenges in maintaining stability and factual grounding in high-stakes clinical scenarios, emphasizing the role of optimization-driven and preference-guided training in reducing hallucination risks and unsafe model behaviors~\cite{singhal2025toward, dai2024intrusion, yang2023potential, yang2023systematic}. 
Related optimization-driven approaches in medical AI, including both language-based and perception-based systems, further demonstrate that structured optimization strategies and customized model designs improve robustness and reliability in safety-critical clinical applications~\cite{li2025mspo, yang2025braincnn, batool2025forensic}.

Building on these advances, we introduce RGPO, a ranking-based extension of preference optimization. RGPO samples multiple responses per prompt and computes relative advantages, thereby reducing variance through group-level aggregation and providing more robust training signals for reasoning-intensive tasks like medical question answering.

\subsection{Ranking-based Optimization in Clinical Reasoning}
Recent studies have expanded preference alignment for LLMs from pairwise to listwise settings, demonstrating that exploiting full ranking information provides denser and more stable supervision than binary preferences~\cite{liu2025lipo, pesaran2025lpoi}.  
Listwise preference optimization frameworks such as Listwise Preference Optimization (LIPO)~\cite{liu2025lipo} show that incorporating graded relevance and positional importance mitigates rank compression and significantly improves alignment performance compared to pairwise-only methods.  

Beyond general alignment, recent work in safety-critical and medical domains further emphasizes the necessity of optimization-driven alignment strategies.  
Med-PaLM~\cite{singhal2025toward} and GatorTron~\cite{yang2022gatortron} demonstrate that preference-guided optimization substantially improves factuality, robustness, and safety in medical reasoning, reducing hallucination risks in high-stakes clinical QA.  

Motivated by these findings, we propose RGPO, which combines listwise ranking with probabilistic refinement and linear position-based rewards to enable robust and clinically aligned optimization for medical reasoning.  
Unlike prior work that focuses on general-domain alignment, our framework explicitly integrates medical-specific reasoning structure, multi-dimensional quality assessment, and rank-aware optimization, achieving strong gains across multiple medical benchmarks.

\begin{figure*}
    \centering
    \includegraphics[width=0.9\textwidth]{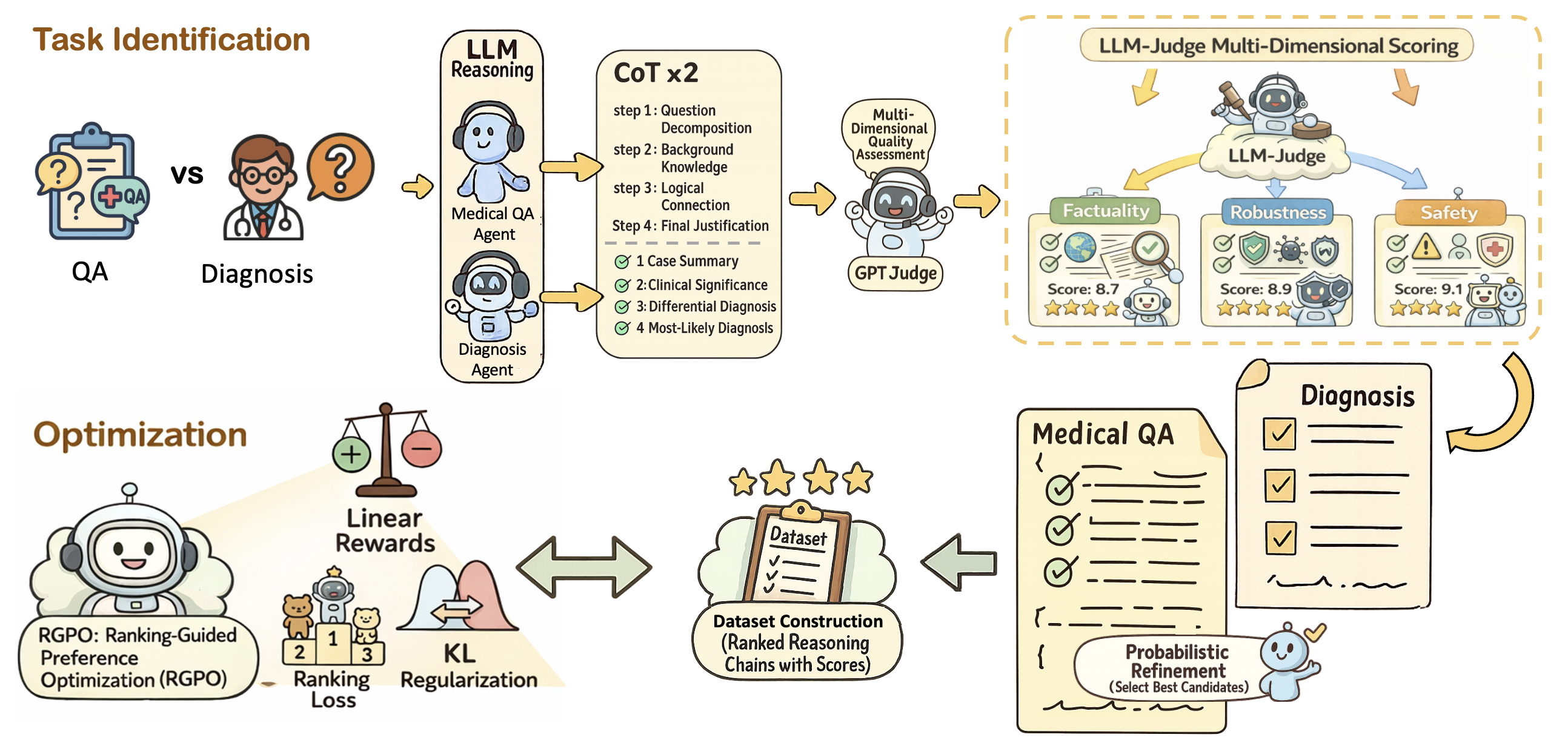}
    \caption{Overview of the proposed pipeline. The framework distinguishes between medical QA and diagnosis tasks, applies chain-of-thought (CoT) reasoning with multi-dimensional quality assessment and probabilistic refinement, and further improves through ranking-guided preference optimization with dataset construction.}
    \label{fig:Fig1}
\end{figure*}

\section{Methodology}
\subsection{Overview}
In this work, we propose a systematic framework, Ranking-Guided Preference Optimization for Reliable Clinical Reasoning (RGPO), to improve medical question answering by constructing and refining Chain-of-Thought (CoT) reasoning data using preference-based ranking signals.
The overall workflow consists of four steps.
(1) Based on task identification, we distinguish between general medical QA and diagnostic reasoning tasks and apply task-adaptive CoT strategies.
(2) For each instance, the CoT is explicitly structured into question decomposition, background knowledge integration, logical reasoning, and final clinical justification, aligning the reasoning process with standard medical workflows.
(3) Each reasoning chain is evaluated using probabilistic refinement rules that capture multiple quality dimensions of medical reasoning.
(4) High-quality reasoning chains are retained and organized into ranked preference groups, while low-quality candidates are filtered based on the refinement scores.
These ranked preference groups provide training signals for RGPO, which integrates listwise preference optimization, linear reward shaping, and Kullback–Leibler (KL)-divergence regularization to guide the model toward reliable medical decision making. Details of the proposed optimization method are provided in Section~IV. 
Terminology and symbol definitions are summarized in Table S2 in the Supplementary Material.

\subsection{Problem Formulation}

We begin by formulating medical questions into two complementary task types to enable domain-appropriate reasoning. We define $\tau_{\text{QA}}$ as general medical question-answering, which targets questions requiring factual medical knowledge and logical reasoning about biomedical concepts, treatments, or underlying mechanisms. Conversely, $\tau_{\text{Diag}}$ represents medical diagnosis tasks involving clinical scenarios that necessitate systematic diagnostic reasoning through detailed patient presentation analysis and comprehensive differential diagnosis consideration. This dual-task framework establishes our task space $\mathcal{T} = \{\tau_{\text{QA}}, \tau_{\text{Diag}}\}$. Let $\mathcal{X}$ and $\mathcal{Q}$ represent the spaces of structured medical records and natural-language prompts, respectively. Each task instance is an input pair $z = (x, q) \in \mathcal{X} \times \mathcal{Q}$, where $x$ is a medical record and $q$ is the corresponding prompt. A task classifier determines the appropriate task $\tau \in \mathcal{T}$ for each instance.

\subsection{Task-Adaptive Chain-of-Thought Generation}\label{TACOTG}
Given an input $z$ and a task label $\tau$, a conditional policy $\pi_\theta(\cdot \mid z,\tau)$ generates $K=5$ candidate Chain-of-Thoughts (CoTs) $C = \{c_k\}_{k=1}^{5}$ for each medical question, where each candidate $c_k$ is sampled as $c_k \sim \pi_\theta(\cdot \mid z,\tau)$. The task label $\tau$ explicitly conditions the generation process, enabling the model to adapt its reasoning structure to different types of medical tasks.

Each candidate reasoning chain $c_k$ is organized into four structured reasoning steps:
\begin{equation}
    c_k = \big(u^{(1)}_k,\, u^{(2)}_k,\, u^{(3)}_k,\, u^{(4)}_k\big),
\end{equation}
where the $j$-th component $u^{(j)}_k$ depends on the task category $\tau$ and is defined as
\begin{equation}
    u^{(j)}_k =
\begin{cases}
s^{(j)}_k, & \text{if } \tau = \tau_{QA}, \\[6pt]
t^{(j)}_k, & \text{if } \tau = \tau_{Diag},
\end{cases}
\qquad j=1,2,3,4.
\end{equation}

For $\tau = \tau_{QA}$, corresponding to general medical question answering, the reasoning chain consists of four components:
\[
\begin{array}{ll}
s^{(1)}: & \text{Question Decomposition},\\
s^{(2)}: & \text{Background Knowledge},\\
s^{(3)}: & \text{Logical Connection},\\
s^{(4)}: & \text{Final Justification.}
\end{array}
\]
This decomposition encourages the model to explicitly separate problem understanding, knowledge retrieval, and logical inference, thereby promoting transparent and verifiable medical reasoning in question answering tasks.

Similarly, for $\tau = \tau_{Diag}$, which focuses on clinical diagnosis, the reasoning components are defined as:
\[
\begin{array}{ll}
t^{(1)}: & \text{Case Summary},\\
t^{(2)}: & \text{Clinical Significance},\\
t^{(3)}: & \text{Differential Diagnosis},\\
t^{(4)}: & \text{Most Likely Diagnosis.}
\end{array}
\]
These steps are designed to mirror standard diagnostic workflows used in clinical practice, guiding the model to progress from symptom summarization to differential reasoning and final diagnostic decision making.

By explicitly aligning each reasoning step with task-specific clinical objectives, the proposed task-adaptive CoT formulation differs from generic Chain-of-Thought prompting, which applies a uniform reasoning structure across tasks without accounting for domain-specific workflows. Thus, the unified representation $c_k = (u^{(1)}_k, u^{(2)}_k, u^{(3)}_k, u^{(4)}_k)$ provides a task-agnostic structural format, while preserving task-dependent semantic specialization through the task label $\tau$.

After generation, the set of $K=5$ candidate reasoning chains is reduced to $M=4$ by retaining the top four candidates based on subsequent quality evaluation and discarding the lowest-ranked one. This step ensures sufficient candidate diversity while limiting the inclusion of low-quality reasoning paths in downstream preference optimization.

\subsection{Medical Prompt Instruction and Reasoning}

Previous research shows that effective prompt template instructions can improve model performance on CoT reasoning~\cite{zhu2024llms, wei2022chain, kojima2022large}. We adopt structured reasoning templates based on established clinical and biomedical reasoning patterns. Table S1 in the Supplementary Material presents example outputs of both the general medical QA reasoning template and the diagnostic reasoning template. The general medical QA template applies a four-step reasoning process, explicitly structuring the model’s reasoning into interpretable stages aligned with clinical logic, and is suitable for general medical question answering tasks such as PubMedQA. Specifically, background knowledge retrieval gathers relevant biomedical facts and principles from medical literature. Logical connection links the retrieved knowledge to the question context. The justification step synthesizes the reasoning to support the final conclusion.

In parallel, our diagnostic reasoning template mirrors the clinical diagnostic workflow used by medical professionals and is designed for diagnostic tasks such as MedQA-USMLE. The process begins with a case summary, which extracts and organizes key clinical findings from patient presentations. Clinical significance analysis then examines symptom patterns and diagnostic indicators to assess their medical relevance. Differential diagnosis systematically considers alternative diagnoses and establishes criteria for elimination. Finally, the most likely diagnosis step concludes with an evidence-supported diagnostic decision based on the available clinical information.

\subsection{Reasoning Refinement Method}

\subsubsection{Multi-Dimensional Quality Assessment Evaluation}
To evaluate reasoning quality across three explicitly defined indicators, coverage, factual accuracy, and redundancy, we propose a joint-probability-based evaluation framework for assessing multiple candidate reasoning paths and selecting the optimal subset from $\mathcal{C} = \{c_1, c_2, \dots, c_n\}$.

The three indicators are assessed by large language models (LLMs) acting as evaluation judges, following established model-as-a-judge and multi-criteria evaluation paradigms for generated text and reasoning quality \cite{liu2023g-eval,kryscinski2020evaluating,es2024ragas,akter2024redundancy, yuan2024self}. Coverage measures whether a reasoning chain sufficiently addresses the key aspects of the medical question or clinical scenario, aligning with notions of completeness and relevance in prior work \cite{liang2022holistic, kryscinski2020evaluating, chang2024survey}. Factual accuracy evaluates whether the reasoning is consistent with established medical knowledge and the provided context, corresponding to factual consistency and faithfulness criteria \cite{manakul2023selfcheckgpt}. Redundancy captures unnecessary repetition or logically overlapping statements within the reasoning chain, reflecting conciseness and information density objectives. Together, these dimensions form a structured, multi-dimensional assessment of reasoning quality beyond coarse binary judgments, enabling more faithful approximation of human evaluation preferences.

For each candidate $c \in \mathcal{C}$, we define three scoring functions $S_i : \mathcal{C} \to \mathcal{S}$ where $i \in \{\text{cov}, \text{fact}, \text{red}\}$, corresponding to coverage $S_{\text{cov}}(c)$, factual accuracy $S_{\text{fact}}(c)$, and redundancy $S_{\text{red}}(c)$. 
Coverage and factual accuracy are evaluated on a 0--5 ordinal scale, where higher scores indicate better quality. Redundancy is scored on a 0--1 scale, where lower scores indicate higher quality (i.e., less redundancy).
These raw scores are mapped into probabilistic indicators $p_i(c) = \pi\big(S_i(c)\big)$ via a normalization mapping $\pi : \mathcal{S} \to [0,1]$.

\subsubsection{Probabilistic Reasoning Refinement}
Unlike traditional additive scoring approaches, our objective is to simultaneously ensure adequate coverage of the source content and maintain factual accuracy, while also suppressing redundancy during generation. In the context of medical reasoning, failure in any single quality dimension may lead to unreliable or unsafe conclusions, making compensatory aggregation across dimensions undesirable. To achieve this, we formulate the decoding objective as the maximization of a multiplicative acceptance function that integrates these three criteria. The multiplicative formulation enforces joint satisfaction of all quality constraints, such that low confidence in any dimension directly reduces the overall acceptance probability and cannot be offset by high scores in other dimensions.

Specifically, the optimal candidate $c^*$ for each question is defined as
\begin{equation}
c^* = \arg\max_{c} \Big[ p_{\text{cov}}(c) \cdot p_{\text{fact}}(c) \cdot \big(1 - p_{\text{red}}(c)\big) \Big],
\end{equation}
where $p_{\text{cov}}(c)$ denotes the probability of coverage, $p_{\text{fact}}(c)$ denotes the probability of factual accuracy, and $p_{\text{red}}(c)$ measures the likelihood of redundancy in the generated content. This design contrasts with weighted or additive scoring schemes, which may allow strong performance in one dimension to mask deficiencies in others.

Next, we introduce the selection operator with the \textit{acceptance function} $P_{\text{accept}}(c)$ as:
\begin{equation}
\Phi_k(\mathcal{C}) = \underset{c \in \mathcal{C}}{\operatorname{Top}\text{-}k} P_{\text{accept}}(c),
\end{equation}
where $\Phi_k(\mathcal{C})$ extracts the top $k$ candidates with the highest acceptance probabilities to be used for subsequent training or decision-making. By unifying multiple quality dimensions into a single acceptance probability, this refinement mechanism provides a principled criterion for ranking and selecting high-quality reasoning chains under preference-based optimization.

Detailed descriptions of the evaluation prompts, scoring methodology, and qualitative examples illustrating these criteria can be found in Sections S2 to S4 of the Supplementary Material.

\section{Training Framework}

\subsection{Ranking-Guided Preference Optimization Algorithm}

In this section, we propose a ranking-guided preference optimization method that differs from traditional approaches such as DPO and Proximal Policy Optimization (PPO)-RLHF, which rely solely on pairwise comparisons (i.e., $c_{1} \succ c_{2}$). Our proposed method leverages complete ranking information (i.e., $c_{1} \succ c_{2} \succ c_{3} \succ \ldots \succ c_{n}$) to optimize model performance. Our goal is to learn a policy $\pi_\theta(\cdot \mid z,\tau)$ that assigns higher likelihood to preferred responses. For each candidate CoT sequence $c_j$ given medical question $z$ and temperature $\tau$, we define the preference score as the conditional log-probability:

\begin{equation}
s_j = \log \pi_\theta(c_j \mid z,\tau) = \frac{1}{|c_j|} \sum_{t=1}^{|c_j|} \log \pi_\theta(c_{j,t} \mid z,\tau, c_{j,<t})
\end{equation}
where $|c_j|$ denotes the sequence length of the CoT, and $c_{j,t}$ represents the token at position $t$ in the reasoning chain. Following the Bradley-Terry model~\cite{bradley1952rank}, the probability that CoT sequence $c_i$ is preferred over sequence $c_j$ is:

\begin{equation}
P(c_i \succ c_j \mid z,\tau) = \sigma\left(\frac{s_i - s_j}{\tau_{BT}}\right) = \frac{1}{1 + \exp\left(-\frac{s_i - s_j}{\tau_{BT}}\right)}
\end{equation}
where $\tau_{BT} > 0$ is a Bradley-Terry temperature parameter (distinct from the task type indicator $\tau$) controlling the sharpness of the preference distribution. Then, we can construct the ranking loss for the $K=5$ candidate CoT sequences extends the Bradley-Terry model to handle complete expert rankings. The $\mathcal{L}_{CoT-rank} $ can be rewritten as:

\begin{equation}
\mathcal{L}_{CoT-rank} = \frac{1}{\binom{K}{2}} \sum_{i=1}^{K-1} \sum_{j=i+1}^{K} \log\left(1 + \exp\left(-\frac{s_i - s_j}{\tau_{BT}}\right)\right).
\end{equation}

\subsection{Linear Rewards From Rank}

Traditional pairwise preference methods (e.g., DPO) 
optimize relative preferences between response pairs but suffer from 
rank compression: they learn correct orderings but assign similar 
scores to top-ranked candidates, limiting the model's ability to 
distinguish the best reasoning chain from merely good ones. Moreover, 
pairwise comparisons discard valuable ranking information when 
evaluators provide complete orderings over multiple candidates.

Group Relative Policy Optimization (GRPO) is one of the widely used algorithms that employs entropy 
regularization to stabilize policy gradient-based learning processes. 
However, it still exhibits sensitivity to parameter tuning, which can 
lead to unstable convergence performance or even divergence during 
training \cite{leahy2022convergence, schulman2015trust, 
mnih2016asynchronous}. To address both the rank 
compression problem and parameter sensitivity, we adopt a 
\textit{linear reward shaping scheme} that 
explicitly leverages the full ranking structure over K candidates.

We then define the advantage value $a_j$ of candidate $c_j$ as:
\begin{equation}
a_j = (K - r_j) - \frac{K-1}{2}, 
\quad j = 1, \dots, K 
\;\;\; \text{s.t.} \;\;
\sum_{j=1}^K a_j = 0.
\label{eq:linear-reward}
\end{equation}
where each candidate is assigned a rank $r_j \in \{1, \dots, K\}$, 
with $r_j = 1$ denoting the highest-quality candidate and $r_j = K$ 
the lowest. This linear transformation enforces that 
rank position matters substantially: being ranked 1st receives much 
higher reward than being ranked 4th, preventing the score compression 
observed in pairwise methods. Intuitively, the linear reward improves 
the model by incentivizing it to shift probability mass from poorly 
ranked reasoning paths toward higher-quality ones. Compared with 
pairwise-only preference optimization, RGPO efficiently incorporates 
full ranking information within each group of candidates, while 
retaining simplicity and stability during optimization.

We then incorporate these linear rewards to increase the probability 
of higher-ranked candidates (positive $a_j$) and decrease the 
probability of lower-ranked ones (negative $a_j$){:}
\begin{equation}
\mathcal{L}_{\text{Linear}} 
= - \frac{1}{K} \sum_{j=1}^K a_j \, s_j.
\label{eq:grpo-loss}
\end{equation}

The linear reward term $\mathcal{L}_{\text{Linear}}$ 
provides global position-based signals across all candidates, while 
$\mathcal{L}_{CoT-rank}$ (Equation~(7)) captures local pairwise 
preference consistency. Their combination (Equation~(13)) addresses 
complementary aspects of preference learning: pairwise methods excel 
at learning relative orderings but suffer from rank compression, while 
position-based rewards enforce the importance of absolute rank but 
lack fine-grained preference structure. By combining both objectives 
through direct summation with equal weighting, we capture both local 
consistency and global ranking structure without requiring additional 
hyperparameter tuning.

\subsection{Training Objective}
To prevent the medical reasoning policy from deviating too far from a 
reference model $\pi_{ref}(\cdot \mid z,\tau)$, we incorporate KL-divergence 
regularization. The sequence-level KL divergence for CoT sequence $c_j$ is:
\begin{equation}
\mathcal{D}_{KL}(c_j \mid z,\tau) = \frac{1}{|c_j|} \sum_{t=1}^{|c_j|} \mathcal{D}_{KL,t}
\end{equation}
The KL regularization term across all $K$ candidates is:
\vspace{-3mm}
\begin{equation}
\mathcal{L}_{KL} = \beta \cdot \frac{1}{K} \sum_{k=1}^{K} \mathcal{D}_{KL}(c_k \mid z,\tau)
\end{equation}
where $\beta \geq 0$ controls the regularization strength for medical reasoning 
stability. Here, the $\mathcal{D}_{KL,t}$ is token-level KL divergence at 
position $t$ in the CoT sequence be expressed as:
\begin{equation}
D_{KL,t} = \sum_{v \in \mathcal{V}} \pi_\theta(v \mid z,\tau, c_{<t}) \log \frac{\pi_\theta(v \mid z,\tau, c_{<t})}{\pi_{ref}(v \mid z,\tau, c_{<t})}
\end{equation}
where $\mathcal{V}$ is the set of all possible tokens in language model. 
During training, we combine the $\mathcal{L}_{CoT-rank}$ and $\mathcal{L}_{KL}$ 
to form $\mathcal{L}_{RGPO-CoT}$ so that the final objective becomes:
\begin{equation}
\mathcal{L}_{total} = \mathcal{L}_{RGPO-CoT} + \mathcal{L}_{\text{Linear}}
\end{equation}

This hybrid design combines ranking-based preference optimization 
($\mathcal{L}_{CoT-rank}$) with linear position rewards ($\mathcal{L}_{\text{Linear}}$) 
to address complementary aspects of preference learning. $\mathcal{L}_{CoT-rank}$ 
optimizes pairwise preference probabilities through the Bradley-Terry model, 
capturing local relative quality differences between reasoning chains (e.g., 
whether candidate A is better than candidate B). However, pairwise methods 
suffer from \textit{rank compression}: they learn correct relative orderings 
but assign similar scores to top-ranked candidates, limiting the model's 
ability to identify the best reasoning chain. In contrast, 
$\mathcal{L}_{\text{Linear}}$ provides global position-based signals across 
all candidates, enforcing that rank position matters substantially (e.g., 
being ranked 1st is much more important than being ranked 4th). We combine 
both objectives through direct summation with equal weighting, as this 
captures both local pairwise consistency and global ranking structure without 
requiring additional hyperparameter tuning.

\section{Experiments}

\subsection{Experimental Setting}

\textbf{Datasets.} We evaluate our approach on two benchmark medical QA datasets. PubMedQA~\cite{jin2019pubmedqa} contains biomedical research questions derived from PubMed abstracts, requiring yes/no/maybe answers based on scientific evidence. MedQA-USMLE~\cite{jin2021disease} consists of multiple-choice diagnostic questions from the United States Medical Licensing Examination, testing clinical reasoning and diagnostic capabilities. These datasets represent complementary aspects of medical reasoning: research-oriented knowledge synthesis and clinical diagnostic decision-making.

In addition, we further evaluate our model on a real-world clinical dataset collected at Far Eastern Memorial Hospital (FEMH), which consists of de-identified electronic medical records and physician diagnostic summaries. Unlike benchmark QA datasets, the FEMH dataset contains free-text clinical presentations and complex multi-symptom cases, providing a more realistic assessment of clinical reasoning performance.

\textbf{Implementation Details.} Our base model is Gemma 2B, chosen for computational efficiency while maintaining competitive performance. We generate five CoT reasoning variants per question and select the four highest-ranking candidates based on their acceptance probabilities for RGPO training. The acceptance threshold $\theta$ is set to 0.6 based on preliminary experiments. All models are trained with AdamW optimizer using a learning rate of 5e-5, batch size of 16, and 3 training epochs. Training is performed on NVIDIA A100 GPUs with mixed precision.

\subsection{Baselines}

We compare against a diverse set of established language models spanning different parameter scales, training strategies, and levels of medical specialization. General-purpose LLMs include Gemma 7B~\cite{team2024gemma}, Mistral 7B~\cite{jiang2023mistral}, Falcon 7B~\cite{almazrouei2023falcon}, Qwen2.5 7B, and Vicuna 7B. These models represent strong modern baselines trained on broad web-scale corpora.

For medical-specialized models, we evaluate MedAlpaca 7B and 13B~\cite{han2023medalpaca}, which are fine-tuned specifically for medical Q\&A tasks, and BioMistral 7B~\cite{labrak2024biomistral}, trained on biomedical literature and clinical text. We additionally include Medicine-LLM, a medical-domain model tailored for clinical reasoning tasks, and Meditron 7B~\cite{chen2024meditron}, a strong medical adaptation of LLaMA trained on PubMed and clinical corpora. Finally, GPT-OSS 20B\cite{agarwal2025gpt} is included as a large open-source frontier model with strong general and biomedical reasoning performance.

Together, these baselines cover a wide spectrum of model sizes (7B–20B), training regimes (general-purpose vs.\ biomedical-pretrained vs.\ medically fine-tuned), and architectural families, offering a comprehensive comparison against both domain-general and domain-specific state-of-the-art systems.

\subsection{Comparison with DPO and SFT}
We compare our RGPO approach against two established training paradigms. Supervised Fine-Tuning (SFT) employs standard next-token prediction on the highest-quality reasoning chains. DPO utilizes pairwise preference learning between high- and low-quality reasoning pairs. RGPO implements group-wise ranking optimization across multiple quality levels, enabling more nuanced preference learning from ranked reasoning chains.

\subsection{Evaluation Metrics}

We use two complementary metrics to assess model performance: accuracy, which measures the proportion of correctly answered questions for direct factual assessment, and macro F1 score, which computes the unweighted average of F1 scores across answer categories to ensure balanced evaluation and address class imbalance. This dual-metric approach captures both overall performance and category-specific reasoning quality.

\begin{table*}
\centering
\caption{Comparison of different models on PubMedQA and MedQA-USMLE under 0-shot, 1-shot, and 5-shot settings. Best results are in \textbf{bold}.
Results for our method (Gemma 2B, RGPO) are averaged over 5 runs with different random seeds.
The standard deviations (Accuracy / Macro-F1) are as follows:
PubMedQA: 0-shot ($\pm$1.41 / $\pm$1.32), 1-shot ($\pm$1.36 / $\pm$1.28), 5-shot ($\pm$1.29 / $\pm$1.35);
MedQA-USMLE: 0-shot ($\pm$1.52 / $\pm$1.47), 1-shot ($\pm$1.24 / $\pm$1.39), 5-shot ($\pm$1.33 / $\pm$1.41).}
\resizebox{\linewidth}{!}{%
\begin{tabular}{l ccc ccc}
\hline
\multirow{2}{*}{\textbf{Models}} &
\multicolumn{3}{c}{\textbf{PubMedQA (Accuracy / Macro F1)}} &
\multicolumn{3}{c}{\textbf{MedQA-USMLE (Accuracy / Macro F1)}} \\
\cline{2-7}
 & \textbf{0-shot} & \textbf{1-shot} & \textbf{5-shot} &
   \textbf{0-shot} & \textbf{1-shot} & \textbf{5-shot} \\
\hline
BioMedLM       & 35.10 / 20.45 & 42.23 / 25.75 & 48.65 / 30.30 & 22.68 / 13.73 & 26.07 / 15.92 & 31.18 / 20.36 \\
BioMistral 7B  & 36.89 / 27.98 & 42.76 / 32.41 & 45.53 / 36.28 & 19.89 / 7.47  & 31.53 / 13.41 & 37.57 / 16.10 \\
Falcon 7B      & 55.83 / 37.72 & 57.14 / 39.08 & 57.92 / 40.63 & 17.23 / 8.14  & 24.87 / 16.21 & 24.89 / 13.94 \\
Gemma 7B       & 58.08 / 48.02 & 58.67 / 49.51 & 58.12 / 50.84 & 20.37 / 10.18 & 43.12 / 24.15 & 51.18 / 32.35 \\
MedAlpaca 7B   & 44.83 / 42.41 & 46.72 / 43.94 & 48.51 / 45.34 & 24.01 / 16.37 & 41.27 / 31.46 & 47.82 / 36.38 \\
Meditron 7B    & 55.32 / 44.82 & 58.65 / 43.50 & 60.44 / 46.27 & 19.23 / 14.77 & 27.88 / 19.94 & 34.01 / 24.33 \\
Medicine-LLM   & 41.87 / 33.76 & 42.25 / 35.91 & 42.18 / 35.58 & 26.35 / 16.90 & 34.37 / 23.46 & 39.82 / 28.47 \\
Mistral 7B     & 51.43 / 47.11 & 54.03 / 49.81 & 56.78 / 52.43 & 22.66 / 11.78 & 40.85 / 23.47 & 49.02 / 29.16 \\
Qwen2.5 7B     & 51.34 / 48.40 & 54.43 / 46.82 & 55.38 / 43.44 & 35.06 / 23.84 & 44.23 / 31.02 & 50.87 / 37.68 \\
Vicuna 7B      & 56.27 / 49.13 & 58.88 / 50.34 & 59.92 / \textbf{53.47} & 27.43 / 17.86 & 33.91 / 22.56 & 40.12 / 27.34 \\

MedAlpaca 13B  & 52.61 / 46.16 & 52.47 / 47.93 & 54.38 / 49.29 & 23.15 / 22.42 & 39.36 / 25.27 & 45.31 / 26.54 \\

GPT-OSS 20B    & 47.27 / 43.75 & 51.23 / 47.95 & 55.61 / 52.34 & 40.02 / 23.76 & \textbf{50.13} / 30.42 & \textbf{52.68} / 32.11 \\
\hline
\textbf{Our Method (Gemma 2B, RGPO)} &
\textbf{60.23} / \textbf{51.78} & \textbf{61.76} / \textbf{50.95} & \textbf{62.02} / 53.12 &
\textbf{42.38} / \textbf{30.51} & 46.27 / \textbf{32.68} & 51.67 / \textbf{35.24} \\
\hline
\end{tabular}%
}
\label{tab:main_results}
\end{table*}

\begin{table*}[h!]
\centering
\caption{
Performance comparison on the FEMH dataset using semantic similarity metrics.
Results for our method are averaged over 5 random seeds. The standard deviations (BERTScore-F1 / Cosine Similarity) are:
0-shot ($\pm$0.016 / $\pm$0.012), 
1-shot ($\pm$0.010 / $\pm$0.015), 
5-shot ($\pm$0.014 / $\pm$0.014).} 

\resizebox{\linewidth}{!}{%
\begin{tabular}{lccc}
\hline
\textbf{Models} & \textbf{0-shot (BERTScore-F1 / Cosine)} & \textbf{1-shot (BERTScore-F1 / Cosine)} & \textbf{5-shot (BERTScore-F1 / Cosine)} \\
\hline
BioMedLM        & 0.798 / 0.342 & 0.812 / 0.351 & 0.825 / 0.353 \\
BioMistral 7B   & 0.830 / 0.398 & 0.842 / 0.326 & 0.885 / 0.405 \\
Falcon 7B       & 0.856 / 0.342 & 0.861 / 0.351 & 0.872 / 0.368 \\
Gemma 7B        & 0.871 / 0.367 & 0.874 / 0.381 & 0.868 / 0.373 \\
MedAlpaca 7B    & 0.848 / 0.334 & 0.852 / 0.343 & 0.861 / 0.357 \\
Mistral 7B      & 0.859 / 0.345 & 0.865 / 0.359 & 0.871 / 0.366 \\
Vicuna 7B       & 0.872 / 0.358 & 0.877 / 0.369 & 0.884 / 0.391 \\
GPT-OSS 20B     & 0.851 / 0.338 & 0.864 / 0.354 & 0.873 / 0.368 \\
\hline
\textbf{Our Method (Gemma 2B, RGPO)} & \textbf{0.879 / 0.423} & \textbf{0.884 / 0.507} & \textbf{0.891 / 0.528}  \\
\hline
\end{tabular}%
}
\label{tab:femh_similarity}
\end{table*}

\begin{table*}[t]
\centering
\caption{
Performance comparison on ICD-10 diagnostic coding using our FEMH dataset. Model-generated diagnoses are mapped to standardized ICD-10 codes for evaluation. Accuracy measures exact code-level agreement, while Recall@k measures whether the correct code appears in the top-k predictions. Results are shown for 0-shot, 1-shot, and 5-shot in-context learning settings.
}
\resizebox{\linewidth}{!}{%
\begin{tabular}{l ccc ccc ccc}
\hline
\multirow{2}{*}{\textbf{Method}} 
& \multicolumn{3}{c}{\textbf{0-shot}} 
& \multicolumn{3}{c}{\textbf{1-shot}} 
& \multicolumn{3}{c}{\textbf{5-shot}} \\
\cline{2-10}
& Accuracy & Recall@3 & Recall@5
& Accuracy & Recall@3 & Recall@5
& Accuracy & Recall@3 & Recall@5 \\
\hline

BioMedLM
& 52.18 & 55.27 & 58.63
& 53.46 & 56.83 & 60.17
& 54.76 & 58.31 & 61.84 \\

BioMistral 7B 
& 53.12 & 56.04 & 59.18
& 54.29 & 57.36 & 59.71
& 55.64 & 58.73 & 61.92 \\

Falcon 7B
& 54.37 & 57.68 & 60.74
& 55.81 & 58.94 & 62.13
& 57.06 & 60.43 & 63.82 \\

Gemma 7B      
& 56.94 & 60.31 & 63.47
& 58.19 & 62.44 & 65.88
& 59.63 & 64.12 & 67.41 \\

MedAlpaca 7B
& 53.82 & 57.03 & 60.11
& 55.21 & 58.52 & 61.83
& 56.63 & 60.11 & 63.72 \\

Mistral 7B    
& 57.71 & 61.63 & 66.04
& 58.79 & 63.24 & 67.39
& 60.11 & 64.72 & 68.81 \\

Vicuna 7B
& 55.27 & 58.33 & 61.74
& 56.44 & 59.73 & 63.14
& 57.86 & 61.31 & 64.72 \\

GPT-OSS 20B   
& 59.74 & 64.38 & 68.11
& 61.02 & 65.71 & 69.43
& 62.21 & 67.11 & 70.92 \\

\hline
\textbf{RGPO (Ours)} 
& \textbf{61.54} & \textbf{65.42} & \textbf{68.15}
& \textbf{66.18} & \textbf{70.67} & \textbf{73.56}
& \textbf{69.53} & \textbf{74.34} & \textbf{76.38} \\

\hline
\end{tabular}%
}
\label{tab:icd_based_evaluation}
\end{table*}

\begin{table}[t]
\centering
\footnotesize
\setlength{\tabcolsep}{4pt}

\caption{Statistical significance of performance improvements over baselines on PubMedQA and MedQA-USMLE across different few-shot settings. Significance levels are determined using McNemar's test for accuracy comparisons.}
\label{tab:appendix_sig_all_models_acc_only}

\resizebox{\columnwidth}{!}{%
\begin{tabular}{l ccc ccc}
\hline
\multirow{2}{*}{\textbf{Baseline Model}} &
\multicolumn{3}{c}{\textbf{PubMedQA}} &
\multicolumn{3}{c}{\textbf{MedQA-USMLE}} \\
\cline{2-7}
& \textbf{0-shot} & \textbf{1-shot} & \textbf{5-shot}
& \textbf{0-shot} & \textbf{1-shot} & \textbf{5-shot} \\
\hline
BioMedLM        & *** & *** & *** & *** & *** & *** \\
BioMistral 7B   & *** & *** & *** & *** & *** & *** \\
Falcon 7B       & *   & *   & *   & *** & *** & *** \\
Gemma 7B        & ns  & ns  & *   & *** & ns  & ns  \\
MedAlpaca 7B    & *** & *** & *** & *** & **  & *   \\
Meditron 7B     & *   & *   & ns  & *** & *** & *** \\
Medicine-LLM    & *** & *** & *** & *** & *** & **  \\
Mistral 7B      & **  & **  & **  & *** & *   & ns  \\
Qwen2.5 7B      & **  & *   & *   & **  & ns  & ns  \\
Vicuna 7B       & *   & ns  & ns  & *** & *** & *** \\
MedAlpaca 13B   & **  & *** & *** & *** & **  & **  \\
GPT-OSS 20B     & *** & *** & *   & *   & ns  & ns  \\
\hline
\multicolumn{7}{l}{\footnotesize ns: $p \geq 0.05$, * $p < 0.05$, ** $p < 0.01$, *** $p < 0.001$} \\
\end{tabular}%
}

\end{table}

\begin{table}
\centering
\caption{Ablation study evaluating the contributions of probabilistic refinement and listwise preference ranking within the RGPO framework on PubMedQA and MedQA-USMLE under zero-shot settings. Results compare baseline SFT, SFT with refinement only, SFT with listwise ranking only, and the full RGPO method. Best results are in \textbf{bold}. }
\resizebox{\linewidth}{!}{%
\begin{tabular}{l c c c c}
\hline
\textbf{Method} & \multicolumn{2}{c}{\textbf{PubMedQA}} & \multicolumn{2}{c}{\textbf{MedQA-USMLE}} \\
& ACC (\%) & Macro F1 (\%) & ACC (\%) & Macro F1 (\%) \\
\hline
SFT                       
& 55.59 & 38.15 & 36.05 & 23.45 \\

SFT + Refinement                                  
& 57.43 & 40.53 & 37.08 & 23.72 \\

SFT + Listwise Ranking     
& 56.09 & 37.41 & 35.18 & 23.47 \\
DPO + Refinement   & 59.90 & 42.72 & 38.26 & 24.35 \\
\hline
\textbf{Our Method: RGPO (Full Method)}               
& \textbf{60.23} & \textbf{51.78} & \textbf{42.38} & \textbf{30.51} \\
\hline
\end{tabular}%
}
\label{tab:ablation_methods}
\end{table}



\begin{table}[t]
\centering
\caption{
Ablation study on the FEMH dataset under clean and noisy (unprocessed) data conditions.
We evaluate the effect of probabilistic refinement filtering under different prompting settings (0-shot, 1-shot, and 5-shot).
``Noisy'' denotes raw clinical records without text cleaning (e.g., line breaks and formatting artifacts).
The acceptance threshold $\theta$ is fixed at 0.6.
Results are reported using semantic similarity metrics (BERTScore-F1 / Cosine Similarity).
Best results within each data condition are shown in \textbf{bold}.
}
\resizebox{\linewidth}{!}{%
\begin{tabular}{l l c c c}
\hline
\textbf{Data} & \textbf{Method} 
& \textbf{0-shot} 
& \textbf{1-shot} 
& \textbf{5-shot} \\
\hline
\multirow{2}{*}{Clean}
& RGPO w/o Filtering
& 0.867 / 0.401
& 0.872 / 0.489
& 0.878 / 0.507 \\

& \textbf{RGPO w/ Filtering}
& \textbf{0.879} / \textbf{0.423}
& \textbf{0.884} / \textbf{0.507}
& \textbf{0.891} / \textbf{0.528} \\

\hline
\multirow{2}{*}{Noisy (w/o Clean)}
& RGPO w/o Filtering
& 0.851 / 0.372
& 0.858 / 0.451
& 0.864 / 0.468 \\

& \textbf{RGPO w/ Filtering}
& \textbf{0.868} / \textbf{0.398}
& \textbf{0.873} / \textbf{0.486}
& \textbf{0.879} / \textbf{0.502} \\
\hline
\end{tabular}%
}
\label{tab:ablation_filtering_clean_noisy}
\end{table}

\begin{table*}[t]
\centering
\scriptsize
\caption{Qualitative comparison of baseline (SFT) and RGPO reasoning on representative cases from the FEMH dataset.}
\label{tab:femh_qual_compare}
\renewcommand{\arraystretch}{1}
\setlength{\tabcolsep}{6pt}

\begin{tabularx}{\textwidth}{@{}X X@{}}

\begin{minipage}[t]{\linewidth}
\noindent\colorbox{black!10}{\parbox{\linewidth}{\textbf{Case 1: Myocarditis (post-Moderna)}}}\\[2pt]

\textbf{\textcolor{blue!60!black}{Input (Symptoms/Objective):}} \\
\textbf{S:} 1101201 Enzyme decrease $\rightarrow$ possible improved myocarditis… 110/11/26 2nd Moderna $\rightarrow$ 11/27 fever $\rightarrow$ 11/29 mild dyspnea, chest tightness (worse when lying) … LMD CPK WNL, TnT 1559 → 210. EKG: mild pericarditis (r/o). CXR: normal heart size.\\
\textbf{O:} BP, HR stable; clear BS; soft abdomen; no edema.\\[4pt]

\textbf{\textcolor{red!70!black}{Baseline (SFT) Reasoning:}}
\begin{tcolorbox}[enhanced, width=\linewidth, boxrule=0.6pt, left=2pt, right=2pt, top=2pt, bottom=2pt]
\textbf{Step 1.} The patient presents with mild dyspnea and chest tightness that worsens when lying down, suggesting cardiopulmonary involvement.\\
\textbf{Step 2.} Laboratory findings show \textcolor{red!70!black}{elevated troponin levels}, indicating \textcolor{red!70!black}{myocardial injury}.\\
\textbf{Step 3.} The combination of \textcolor{red!70!black}{symptoms and biomarker elevation} is consistent with \textcolor{red!70!black}{myocarditis}.\\
\textbf{Step 4.} Given the clinical presentation, the most likely diagnosis is \textcolor{red!70!black}{acute or post-infectious myocarditis}.\\[2pt]
\textbf{Diagnosis:} \textcolor{red!70!black}{Post-infectious myocarditis}.
\end{tcolorbox}

\textbf{\textcolor{green!40!black}{Model Reasoning (RGPO):}}
\begin{tcolorbox}[enhanced, width=\linewidth, boxrule=0.6pt, left=2pt, right=2pt, top=2pt, bottom=2pt]
\textbf{Step 1.} Identify the key symptoms: The patient presents with fever, dyspnea, chest tightness, and mild discomfort.\\
\textbf{Step 2.} Consider the laboratory results: The patient has \textcolor{green!40!black}{elevated troponin levels, but CK-MB is normal}. \textcolor{green!40!black}{This suggests a possible myocarditis}.\\
\textbf{Step 3.} Integrate the clinical findings: The patient's symptoms and laboratory results suggest a possible myocarditis. The \textcolor{green!40!black}{EKG is normal, and the CXR is normal}.\\
\textbf{Step 4.} Consider differential diagnoses: The differential diagnoses include \textcolor{green!40!black}{myocarditis, pericarditis, and other causes of chest pain and dyspnea}.\\[2pt]
\textbf{Diagnosis:} \textcolor{green!40!black}{Myocarditis}.
\end{tcolorbox}

\textbf{\textcolor{orange!80!black}{Prediction \& Answer:}}\\
Baseline Prediction: Post-infectious myocarditis.\\
Our Prediction: Myocarditis.\\
True Diagnosis: r/o mild myocarditis.
\end{minipage}

&

\begin{minipage}[t]{\linewidth}
\noindent\colorbox{black!10}{\parbox{\linewidth}{\textbf{Case 2: IOL Dislocation (OD)}}}\\[2pt]

\textbf{\textcolor{blue!60!black}{Input (Symptoms/Objective):}}\\
\textbf{S:} Stinging sensation (2021/11), high IOP ($>$50), blurred vision OD. Lost f/u 2 years.\\
\textbf{O:} IOP: 53→17 (treated). Cornea clear. PCIOL subluxation inferiorly. OS: NS 1+–2+. Fundus: C/D 0.5–0.6 (OD). OCT RNFL OD 32.\\[4pt]

\textbf{\textcolor{red!70!black}{Baseline (SFT) Reasoning:}}
\begin{tcolorbox}[enhanced, width=\linewidth, boxrule=0.6pt, left=2pt, right=2pt, top=2pt, bottom=2pt]
\textbf{Step 1.} The patient presents with a stinging sensation in the eye, \textcolor{red!70!black}{markedly elevated intraocular pressure, and blurred vision}, indicating an ocular pathology affecting visual acuity.\\
\textbf{Step 2.} The symptoms show a temporal progression, with stinging sensation noted in late 2021 and blurred vision reported since early 2019.\\
\textbf{Step 3.} Relevant ocular history includes \textcolor{red!70!black}{intraocular lens subluxation}, corneal abrasion, and a prior cerebrovascular accident, suggesting chronic postoperative or structural ocular changes.\\
\textbf{Step 4.} Integrating the symptoms, temporal course, and ocular history, the findings are interpreted as most consistent with \textcolor{red!70!black}{posterior capsular opacification}.\\[2pt]
\textbf{Diagnosis:} \textcolor{red!70!black}{Posterior capsular opacification (PCO)}.
\end{tcolorbox}

\textbf{\textcolor{green!40!black}{Model Reasoning (RGPO):}}
\begin{tcolorbox}[enhanced, width=\linewidth, boxrule=0.6pt, left=2pt, right=2pt, top=2pt, bottom=2pt]
\textbf{Step 1.} Potential causes: \textcolor{green!40!black}{corneal opacity, cataract, glaucoma, IOL subluxation}.\\
\textbf{Step 2.} History: \textcolor{green!40!black}{high IOP, decreased vision, recent IOL subluxation}.\\
\textbf{Step 3.} Comprehensive exam to assess cornea/lens/retina.\\
\textbf{Step 4.} Conclusion: likely \textcolor{green!40!black}{IOL subluxation related to prior surgery/high IOP}.\\[2pt]
\textbf{Diagnosis:} \textcolor{green!40!black}{IOL subluxation}.
\end{tcolorbox}

\textbf{\textcolor{orange!80!black}{Prediction \& Answer:}}\\
Baseline Prediction: Posterior capsular opacification (PCO).\\
Our Prediction: IOL subluxation.\\
True Diagnosis: IOL dislocation (OD).
\end{minipage}

\\
\end{tabularx}
\end{table*}

\begin{table*}[t]
\centering
\caption{Qualitative comparison of reasoning generated by standard SFT and RGPO on representative clinical cases, illustrating common reasoning limitations and improved clinical alignment.}
\label{tab:femh_qual_compare_diff}
\begin{tabular}{p{2.5cm} p{4cm} p{4.5cm} p{4.5cm}}
\toprule
Case & Question (summary) & SFT reasoning (baseline) & RGPO reasoning \\
\midrule
Case 1 &
Post-vaccine myocarditis &
Primarily relies on symptom--biomarker association to support myocarditis, with limited exploration of alternative diagnoses or explicit articulation of residual diagnostic uncertainty. &
Integrates symptoms, laboratory trends, and imaging findings while explicitly considering differential diagnoses and articulating diagnostic uncertainty consistent with clinical practice. \\
\midrule
Case 2 &
IOL dislocation &
Attributes symptoms to posterior capsular opacification (PCO) despite documented intraocular lens subluxation and markedly elevated intraocular pressure, underweighting relevant structural findings. &
Prioritizes objective structural abnormalities, systematically evaluates plausible causes, and appropriately links symptoms to intraocular lens subluxation associated with prior surgery and high IOP. \\
\bottomrule
\end{tabular}
\end{table*}

\textbf{Main Results.} 
Table~\ref{tab:main_results} presents the performance comparison across all evaluated models under 0-shot, 1-shot, and 5-shot settings on both datasets. 
Our method consistently achieves strong gains across all configurations. 
On \textbf{PubMedQA}, RGPO (Gemma 2B) attains 60.23\%, 61.76\%, and 62.02\% accuracy, with corresponding Macro F1 scores of 51.78\%, 50.95\%, and 53.12\% in the 0-shot, 1-shot, and 5-shot settings, respectively. 
Compared to the best baseline (Gemma 7B), this represents improvements of up to 3.9\% in accuracy and 3.8\% in Macro F1, demonstrating the robustness of our refinement mechanism across different few-shot conditions. 
On \textbf{MedQA-USMLE}, our model achieves 42.38\%, 46.27\%, and 51.67\% accuracy, with Macro F1 scores of 30.51\%, 32.68\%, and 35.24\%. 
While maintaining competitive accuracy, RGPO consistently delivers higher Macro F1 than the strongest baseline (GPT-OSS 20B), with gains of +6.75\%, +2.26\%, and +3.13\% across the three settings, highlighting better reasoning completeness and answer quality. 
Overall, our 2B-parameter model matches or surpasses much larger 7B–20B models, demonstrating the effectiveness and efficiency of our reasoning refinement strategy.

\textbf{Statistical Significance and Robustness.}
To support claims of superiority, we conduct paired significance testing on per-question correctness.
For Accuracy, we employ McNemar's test (Table~\ref{tab:appendix_sig_all_models_acc_only}), where each baseline is compared against our method on the identical evaluation set; bold entries indicate statistically significant differences ($p < 0.05$).
In addition, we report the mean and standard deviation over five random seeds for our method in Table~\ref{tab:main_results} for both Accuracy and Macro-F1, reflecting variability across independent training runs.
We note that McNemar's test is specifically applicable to binary correctness and therefore directly supports Accuracy comparisons.
Improvements in Macro-F1 are interpreted in conjunction with consistency across random seeds, as well as the accompanying ablation and error analyses, providing complementary evidence of robustness.

\textbf{Why can a 2B model outperform larger models?}
We attribute this effect primarily to differences in the training procedure rather than model scale alone.
Ranked preference optimization exploits the relative ordering among multiple reasoning candidates, yielding a denser and more structured supervision signal than single-chain supervision or pairwise preference optimization.
This effect is further strengthened by probabilistic refinement, which suppresses low-quality or noisy reasoning traces and improves the effective signal-to-noise ratio of the training data.
Together, these mechanisms reshape the optimization landscape faced by the model and enable consistent improvements in reasoning quality for the 2B model, as corroborated by the ablation studies and qualitative analyses presented below.

\textbf{Clinical Evaluation on FEMH Dataset.}
To further verify the generalization and clinical applicability of our approach,
We conducted an evaluation on the real-world clinical dataset from FEMH.
Each case consists of de-identified electronic medical records containing a clinician-written
subjective description (S) and objective examination findings (O), along with the ground-truth
diagnostic assessment (A). The model is required to generate the diagnostic statement based solely on the S+O inputs,
and the generated assessment is compared against the physician-annotated ground truth.

We evaluate the semantic similarity between the model-generated diagnosis and the true assessment using BERTScore-F1 and sentence-level Cosine Similarity under 0-shot, 1-shot, and 5-shot settings. 
Unlike multiple-choice QA benchmarks (PubMedQA, MedQA-USMLE), the FEMH dataset contains free-text diagnostic statements without predefined answer options, making exact-match accuracy insufficient. The same diagnosis can be expressed using different clinically equivalent terminology (e.g., ``myocarditis,'' ``inflammation of the myocardium''), and strict matching would incorrectly penalize such cases. We therefore adopt BERTScore-F1 and Cosine Similarity to capture semantic and clinical meaning consistency in real-world diagnosis generation. 
To further provide clinically grounded discrete evaluation, we additionally include an International Classification of Diseases (ICD)-10 code-based evaluation (Table~\ref{tab:icd_based_evaluation}), where model outputs are mapped to standardized diagnosis codes and evaluated using Accuracy and Recall@k. Together, these evaluations provide complementary views of clinical reasoning quality and diagnostic correctness.

As shown in Table~\ref{tab:femh_similarity}, our proposed RGPO (Gemma 2B) achieves the best performance across all configurations, obtaining BERTScore-F1 / Cosine scores of 0.879 / 0.423, 0.884 / 0.507, and 0.891 / 0.528 in the respective settings. 
Compared with larger medical-tuned models such as BioMistral-7B and MedAlpaca-7B, RGPO demonstrates higher semantic coherence and diagnostic consistency, highlighting strong few-shot generalization and reliable clinical reasoning despite its compact model size.

To evaluate clinical diagnostic correctness beyond semantic similarity, we conduct ICD-10-based diagnostic agreement analysis, as shown in Table~\ref{tab:icd_based_evaluation}. 
Model-generated diagnoses are mapped to standardized ICD-10 codes and compared against physician-annotated ground truth. 
This approach measures exact diagnostic agreement at the code level, eliminating ambiguity from terminological variations.
As demonstrated in Table~\ref{tab:icd_based_evaluation}, RGPO achieves the highest performance across all metrics (Accuracy, Recall@3, and Recall@5) in all few-shot settings, demonstrating that our method not only generates semantically coherent text but also maintains superior clinical diagnostic accuracy.

Representative reasoning examples generated by our method on the FEMH dataset are shown in Table~\ref{tab:femh_qual_compare}. These cases illustrate how the model integrates subjective symptoms and objective findings to produce structured diagnostic reasoning that aligns closely with clinician-provided assessments.
In Case 1, a post-Moderna vaccination myocarditis case, the baseline SFT model focuses primarily on symptom-biomarker associations and produces post-infectious myocarditis'' without acknowledging the temporal relationship with recent vaccination. In contrast, RGPO explicitly enumerates differential diagnoses (myocarditis, pericarditis, and other cardiopulmonary causes), systematically integrates the vaccination timeline and troponin trends, and distinguishes between Creatine Kinase-MB (CK-MB) and troponin findings. The model's prediction of myocarditis'' more closely matches the ground truth (``r/o mild myocarditis''), demonstrating improved diagnostic precision.
Case 2 further illustrates these benefits. The baseline model incorrectly attributes symptoms to posterior capsular opacification (PCO) despite documented Posterior Chamber Intraocular Lens (PCIOL) subluxation and elevated Intraocular Pressure (IOP) in the objective findings. This suggests that baseline reasoning underweights objective structural abnormalities. RGPO systematically evaluates multiple causes (corneal opacity, cataract, glaucoma, IOL subluxation) and correctly identifies Intraocular Lens (IOL) subluxation,'' which aligns with the true diagnosis of IOL dislocation (OD).''
Table~\ref{tab:femh_qual_compare_diff} summarizes these differences. Across both cases, RGPO demonstrates (1) systematic enumeration of differential diagnoses, (2) explicit integration of temporal and structural findings, and (3) evidence-based justification mirroring clinical workflows. These improvements suggest that ranked preference optimization combined with probabilistic refinement successfully guides the model toward more thorough, interpretable, and clinically aligned reasoning.

\textbf{Ablation Studies.} 
We conduct a comprehensive set of ablation experiments to assess how different components and key hyperparameters influence the behavior of RGPO. In addition to examining the effects of various training objectives and refinement strategies, we focus on three critical hyperparameters: the optimization weight $\beta$, which controls the balance between supervised alignment and preference-driven updates; the rollout budget $K$, which determines the diversity and stability of sampled reasoning trajectories; and the acceptance threshold, which governs the selectivity of the refinement process. By systematically varying these factors across the PubMedQA, MedQA-USMLE, and FEMH dataset, we evaluate their impact on convergence behavior, reasoning quality, and overall task performance.
Overall, these ablations are designed to disentangle the relative contributions of ranking-based optimization and probabilistic refinement, as well as to identify the primary drivers of the observed performance gains.

\smallskip
\noindent
\textbf{Ablation on Training Objectives and Refinement.}
We begin our analysis by examining how different training objectives and 
refinement strategies contribute to overall performance. As shown in 
Table~\ref{tab:ablation_methods}, standard supervised fine-tuning (SFT) 
with refinement provides a strong baseline, achieving 57.43\% 
Accuracy and 40.53\% Macro-F1 on PubMedQA. However, adding listwise ranking 
alone (Row~3: SFT + Listwise Ranking) yields minimal improvement and even 
slightly decreases Macro-F1 (37.41\% vs. 38.15\% for the SFT baseline). This 
reveals a critical insight: ranking optimization over noisy candidates provides 
weak training signals. Without refinement to filter erroneous reasoning, the 
model learns to rank poorly-reasoned outputs, which does not improve accuracy.

When we incorporate both refinement and pairwise preference 
optimization (DPO + Refinement, Row~4), performance improves to 59.90\% / 42.72\%, 
demonstrating the value of combining data quality filtering with preference 
learning. However, the most substantial gain occurs when adding the linear 
reward component to form the full RGPO method (Row~5): 60.23\% / 51.78\%, 
representing a \textbf{+9.06\% improvement in Macro-F1} over DPO+Refinement.

This +9.06\% gain directly validates our hybrid loss design. 
DPO+Refinement already implements strong pairwise preference optimization over 
high-quality candidates, yet it achieves only 42.72\% Macro-F1. The addition 
of $\mathcal{L}_{\text{Linear}}$ yields an additional 9-point gain, confirming 
that linear position rewards address rank compression, a fundamental limitation 
that pairwise methods cannot resolve. Pairwise optimization learns correct 
relative orderings but does not distinguish position importance, leading to 
similar scores across top-ranked items. $\mathcal{L}_{\text{Linear}}$ explicitly 
corrects this by forcing score separation based on ranking position.

These results demonstrate that: (1)~refinement curates a high-quality 
candidate pool, (2)~$\mathcal{L}_{CoT-rank}$ learns preference structure within 
this clean data, and (3)~$\mathcal{L}_{\text{Linear}}$ prevents compression and 
significantly improves top-1 reasoning quality. Together, these components address 
complementary failure modes rather than providing redundant supervision.
Notably, ranking-based preference optimization accounts for the majority of 
the performance gain, while probabilistic refinement primarily improves stability 
and consistency, indicating that the two components play complementary but 
distinct roles. This motivates the more fine-grained hyperparameter ablations 
presented in the following subsections.

\begin{figure*}
    \centering
    \includegraphics[width=1\textwidth]{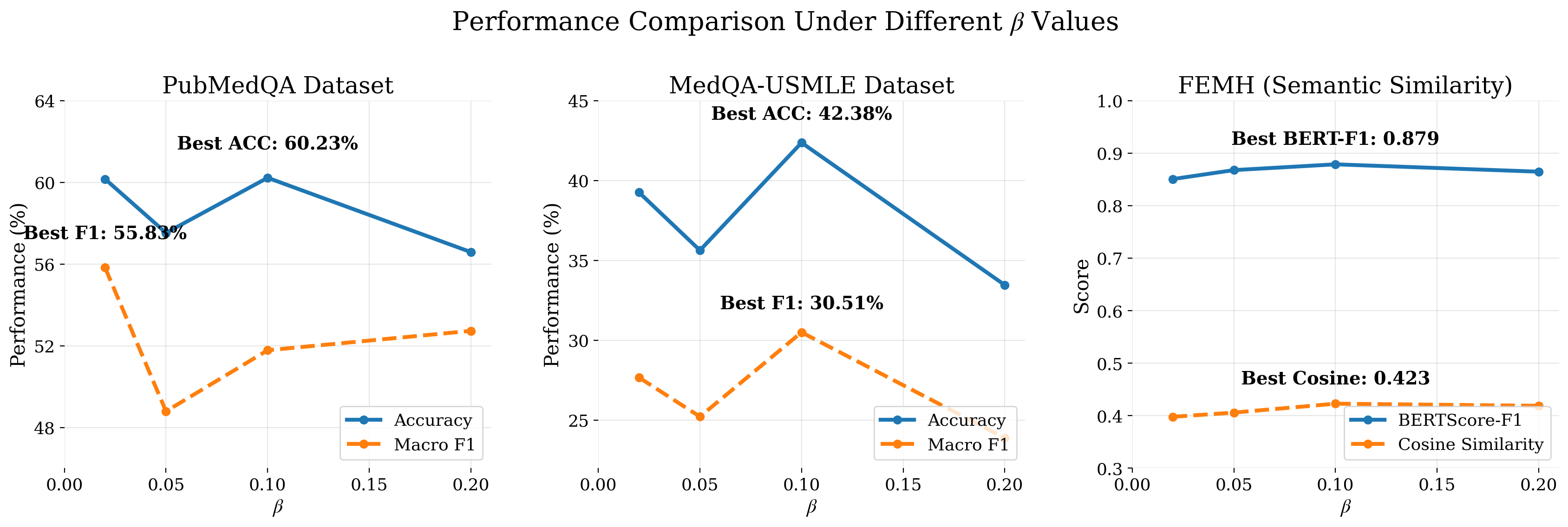}
    \caption{Performance comparison on PubMedQA, MedQA-USMLE, and FEMH dataset under different $\beta$ values. 
PubMedQA/MedQA report Accuracy (solid blue) and Macro F1 (dashed orange), 
while FEMH dataset reports BERTScore-F1 (solid blue) and Cosine Similarity (dashed orange).
}
    \label{fig:Fig3}
\end{figure*}

\noindent
\textbf{Ablation on $\boldsymbol{\beta}$ Values.}
Fig.~\ref{fig:Fig3} presents the performance comparison under different $\beta$ values on the PubMedQA and MedQA-USMLE datasets. 
We vary $\beta \in \{0.0, 0.05, 0.1, 0.2\}$ to control the balance between supervised alignment and preference-driven optimization. 
As shown in the figure, moderate $\beta$ values (around 0.1) yield the best results, achieving the highest Accuracy (60.23\% on MedQA-USMLE, 42.38\% on PubMedQA) and Macro~F1 (55.83\% and 30.51\%, respectively). 
When $\beta$ is too small, the model underfits the preference signal; when $\beta$ is too large, over-regularization reduces reasoning diversity. 
This indicates that an appropriate trade-off between the base objective and preference regularization is crucial for stable optimization.

A similar trend is observed on the FEMH dataset (right panel of Fig.~\ref{fig:Fig3}), where $\beta = 0.1$ achieves the best semantic alignment, with a BERTScore-F1 of 0.879 and a cosine similarity of 0.423. 
Both metrics degrade when $\beta$ deviates from this value, further confirming that moderate preference regularization leads to the most coherent and clinically meaningful diagnostic reasoning.

\begin{figure*}
    \centering
    \includegraphics[width=1\textwidth]{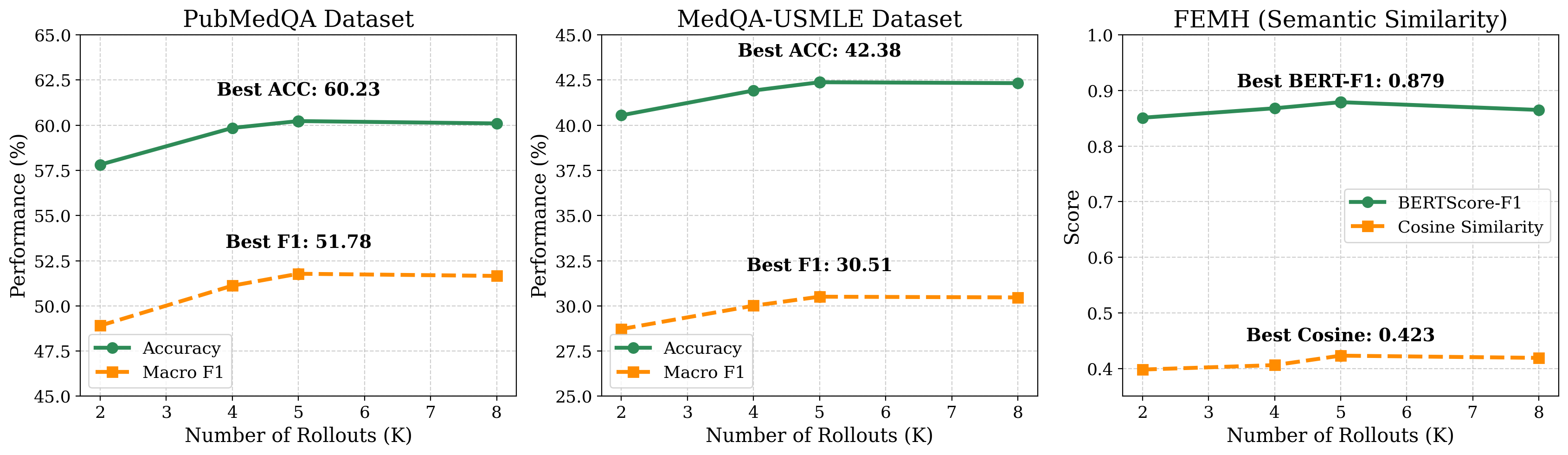}
    \caption{
    Performance comparison on PubMedQA, MedQA-USMLE, and FEMH dataset under different rollout numbers ($K$). 
PubMedQA/MedQA report Accuracy and Macro F1; FEMH dataset reports BERTScore-F1 and Cosine Similarity. 
Each prompt generates $K$ candidate CoTs. For $K<4$, all candidates are used ($M=K$); for $K\ge4$, the top $M=4$ candidates are selected for training.
    }
    \label{fig:Fig4}
\end{figure*}

\smallskip
\noindent
\textbf{Ablation on Rollout Budget $\boldsymbol{K}$.}
To further investigate the influence of rollout budget on reasoning quality, we vary the number of rollouts $K \in {2, 4, 5, 8}$ during preference optimization. As illustrated in Fig.~\ref{fig:Fig4}, increasing $K$ from 2 to 5 leads to steady improvements in both Accuracy and Macro~F1 on PubMedQA and MedQA-USMLE. However, when $K$ exceeds 5, performance gains saturate and even slightly decline, indicating that excessive rollouts introduce redundant samples and reward noise, increasing computational cost without meaningful benefit. Empirically, $K{=}5$ provides the best trade-off between performance and efficiency, consistent with prior observations in RLHF and DPO frameworks where higher rollout counts yield diminishing returns.

On the FEMH dataset (right panel of Fig.~\ref{fig:Fig4}), a similar trend is observed. Semantic similarity metrics including BERTScore-F1 and cosine similarity increase as $K$ grows to 4 or 5, reaching peak values of 0.879 and 0.423 (shown as 0.88 and 0.42 in the figure due to rounding). When $K$ is further increased, both metrics slightly degrade, suggesting that excessive rollouts introduce noisy or redundant chains of thought and weaken alignment. These results confirm that a moderate rollout budget yields the most coherent and clinically meaningful diagnostic reasoning.

\begin{figure*}
    \centering
    \includegraphics[width=1\textwidth]{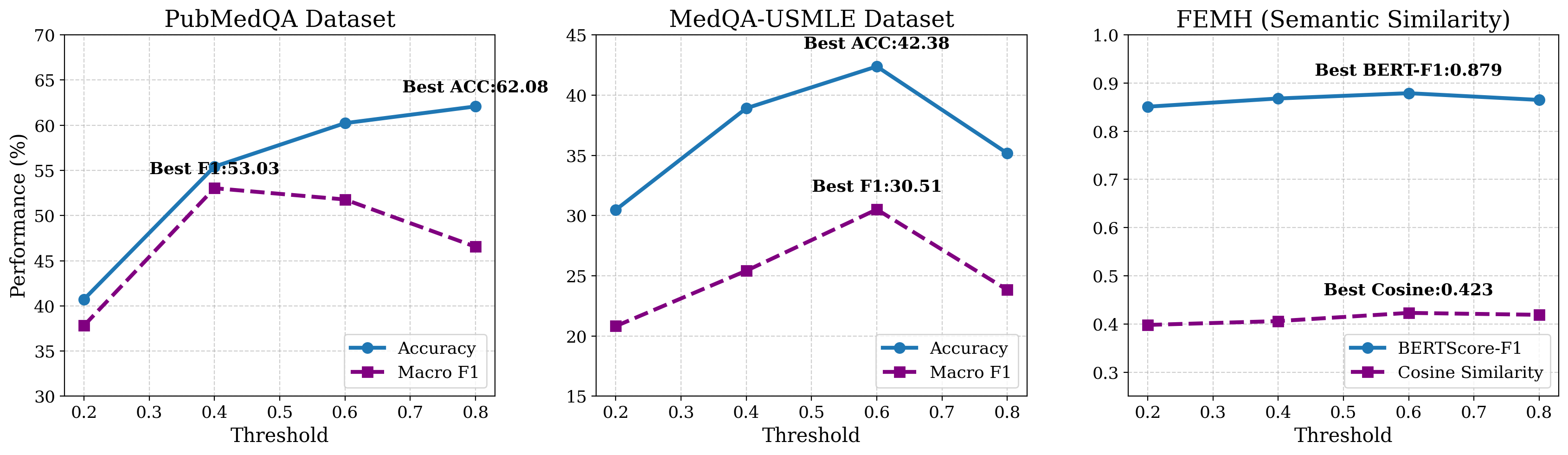}
    \caption{Performance across different acceptance thresholds on PubMedQA, MedQA-USMLE, and FEMH dataset. 
PubMedQA/MedQA report Accuracy (solid blue) and Macro F1 (dashed purple), while FEMH dataset reports BERTScore-F1 and Cosine Similarity.
}
    \label{fig:threshold_ablation}
\end{figure*}

\textbf{Threshold and Convergency Analysis.} 
Fig.~\ref{fig:threshold_ablation} examines the impact of the acceptance probability threshold $\theta$ on model performance.
The threshold $\theta$ serves as an acceptance-based filtering gate that determines which candidate reasoning chains are allowed to enter subsequent preference optimization.
Lower thresholds (0.2, 0.4) result in excessive refinement, potentially degrading reasoning quality through over-correction. The optimal threshold of 0.6 balances the necessity for refinement with the preservation of reasoning. Higher thresholds (0.8) show marginal accuracy improvements but decreased Macro F1, suggesting that overly conservative refinement may miss opportunities for quality enhancement while maintaining correctness.
This degradation at lower thresholds indicates that setting $\theta$ too low allows a large number of low-quality or misleading reasoning chains to pass the acceptance-based filtering stage and enter preference optimization, thereby introducing noisy preference signals during training. Based on this observation, we adopt a conservative threshold of $\theta = 0.6$ in our main experiments to balance refinement effectiveness and training stability.

\begin{figure*}
    \centering
    \includegraphics[width=1\textwidth]{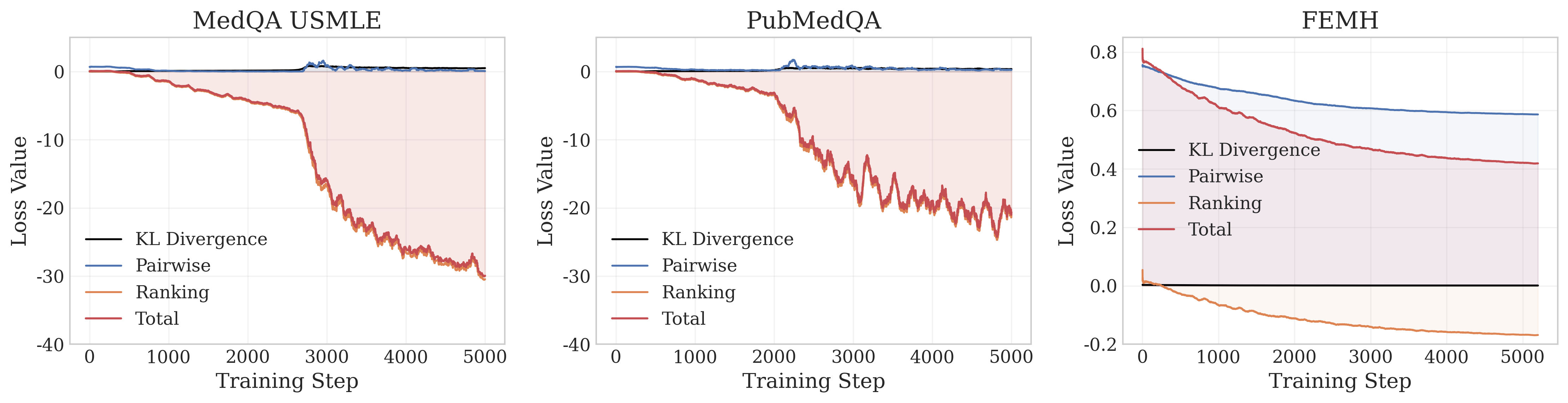}
    \caption{Training loss curves on MedQA-USMLE, PubMedQA, and FEMH dataset. 
The plots show KL divergence, pairwise loss, ranking loss, and total loss across training steps for each dataset.
}
    \label{fig:loss_plot}
\end{figure*}

To evaluate the robustness of our training procedure, we further analyze the loss dynamics of RGPO as shown in Fig.~\ref{fig:loss_plot}. The results demonstrate that RGPO exhibits stable and convergent loss curves on both MedQA-USMLE and PubMedQA, confirming that the ranking-based preference optimization framework maintains reliable optimization behavior throughout training. The group-level ranking objective also effectively mitigates the high-variance fluctuations typically observed in pairwise preference learning.

On the FEMH dataset (right panel of Fig.~\ref{fig:loss_plot}), we observe a similarly stable convergence pattern. Both the pairwise and ranking components decrease smoothly over training steps, and the total loss follows a consistent downward trajectory without oscillatory behavior. This indicates that RGPO also achieves stable optimization when applied to real-world clinical narratives, further validating its robustness and applicability beyond benchmark QA datasets.

\textbf{Effect of Refinement Filtering.}
We further analyze the impact of probabilistic refinement filtering on the FEMH dataset.
Table~\ref{tab:ablation_filtering_clean_noisy} compares RGPO with probabilistic refinement \emph{with} and \emph{without} acceptance-based filtering under different prompting settings.
Across all shot configurations, applying refinement filtering consistently improves semantic similarity scores, indicating that filtering low-confidence reasoning chains before preference optimization is crucial for stable and effective training.

\textbf{Reasoning Results from PubMedQA and MedQA-USMLE Tasks.}
The reasoning examples shown in Table S1 and Fig. S3 in the Supplementary Material provide representative cases generated by our model using the proposed prompt templates.
In PubMedQA, the model explicitly decomposes the question, incorporates relevant biomedical background knowledge, and connects evidence to the final answer, reflecting a literature-grounded reasoning pattern.
In MedQA-USMLE, the reasoning traces follow a clinical workflow, including case summarization, identification of key findings, differential diagnosis, and treatment selection, demonstrating the model's ability to mimic expert diagnostic procedures.
Across both tasks, the generated reasoning showcases coherent multi-step inference, consistent structure, and appropriate use of domain knowledge.
Together, these examples show that our approach yields structured and interpretable reasoning that is well aligned with expert biomedical and clinical decision-making, offering qualitative evidence of the model’s improved reasoning quality.

\section{Conclusion}

This paper introduces \textbf{RGPO}, a ranking-guided preference optimization framework for enhancing medical reasoning in large language models through structured preference learning and probabilistic reasoning refinement. By combining listwise preference optimization with probabilistic quality assessment, RGPO promotes clinically coherent and evidence-consistent reasoning beyond surface-level pattern matching.

Experiments on PubMedQA, MedQA-USMLE, and the FEMH dataset demonstrate that a compact 2B-parameter model trained with RGPO achieves consistent and statistically significant improvements over both general-purpose and medically fine-tuned baselines. Ablation studies further indicate that these gains arise from the complementary interaction between ranked preference optimization and probabilistic refinement, rather than from model scaling alone.

These results highlight the effectiveness of preference-guided optimization and structured refinement as scalable strategies for improving clinical reasoning quality in resource-efficient medical AI systems.

\section{Limitations and Future Work}

Despite its effectiveness, RGPO has several important limitations.

First, the probabilistic refinement mechanism relies on a learned judge model that emphasizes confidence calibration rather than external factual verification. As a result, internally coherent but factually incorrect reasoning may occasionally pass the filter, reflecting the inherent difficulty of distinguishing well-structured but incorrect reasoning without external knowledge grounding.

Second, the refinement process operates in a largely closed-loop manner, where model-generated reasoning is evaluated by another model with similar parametric biases. This introduces the risk of reinforcing high-confidence errors, particularly in safety-critical medical scenarios, due to the absence of independent verification mechanisms.

Third, our evaluation is primarily based on benchmark datasets and retrospective clinical cases, which may not fully capture the ambiguity, noise, and distribution shifts characteristic of real-world clinical inputs.

Future work will address these limitations by extending RGPO to additional medical reasoning benchmarks and multimodal clinical tasks, and by integrating external medical knowledge through retrieval-augmented verification frameworks. We will further explore hybrid validation architectures combining LLM-based judges with symbolic or rule-based medical validators, as well as clinician-in-the-loop evaluation protocols for high-risk refinement stages.

These directions aim to transform RGPO from a calibration-based refinement framework into a verification-aware, externally grounded system suitable for real-world clinical deployment. Human-centered evaluations with clinical experts will be essential for ensuring safety, interpretability, and deployment reliability.


\section{Ethical approval}

The study was approved by the hospital's Research Ethics Review Committee, and all data were fully de-identified. All ethics review procedures and data collection were conducted in accordance with the Committee's standard guidelines and regulations.


\section*{Supporting Information}

This work was supported by the Far Eastern Memorial Hospital Innovation Research Project (Grant No. FEMH-2023-007).


\bibliographystyle{IEEEtran}
\bibliography{sample}

\end{document}